\documentclass[10pt,twocolumn,letterpaper]{article}

\usepackage[pagenumbers]{cvpr} %

\usepackage{graphicx}
\usepackage{amsmath}
\usepackage{amssymb}
\usepackage{booktabs}
\usepackage[numbers,sort]{natbib}
\usepackage{siunitx}

\usepackage[pagebackref,breaklinks,colorlinks]{hyperref}

\def\cvprPaperID{9925} %
\def\confName{CVPR}
\def\confYear{2022}

\newcommand{\G}{\mathcal{G}}

\renewcommand{\P}{\mathbf{P}}

\newcommand{\I}{\mathcal{I}}

\newcommand{\f}{\mathbf{f}}

\renewcommand{\l}{\mathbf{l}}

\newcommand{\p}{\mathbf{p}}

\renewcommand{\u}{\mathbf{u}}

\renewcommand{\v}{\mathbf{v}}

\newcommand{\w}{\mathbf{w}}

\newcommand{\nop}[1]{}

\def\eg{\emph{e.g.}}
\def\ie{\emph{i.e.}}

\begin{document}

\title{Neural Face Identification in a 2D Wireframe Projection of a Manifold Object}

\author{
Kehan Wang\thanks{Work done during an internship at Manycore Tech Inc.}
\\
University of California, Berkeley \\
{\tt\small wang.kehan@berkeley.edu} \quad
\and 
Jia Zheng \quad
Zihan Zhou\\
Manycore Tech Inc.\\
{\tt\small \{jiajia, shuer\}@qunhemail.com}
}

\maketitle

\begin{abstract}
In computer-aided design (CAD) systems, 2D line drawings are commonly used to illustrate 3D object designs. To reconstruct the 3D models depicted by a single 2D line drawing, an important key is finding the edge loops in the line drawing which correspond to the actual faces of the 3D object. In this paper, we approach the classical problem of face identification from a novel data-driven point of view. We cast it as a sequence generation problem: starting from an arbitrary edge, we adopt a variant of the popular Transformer model to predict the edges associated with the same face in a natural order. This allows us to avoid searching the space of all possible edge loops with various hand-crafted rules and heuristics as most existing methods do, deal with challenging cases such as curved surfaces and nested edge loops, and leverage additional cues such as face types. We further discuss how possibly imperfect predictions can be used for 3D object reconstruction. The project page is at \url{https://manycore-research.github.io/faceformer}.
\end{abstract}

\section{Introduction}

In this paper, we revisit a classical problem in computer-aided design (CAD), namely the conversion of 2D line drawings into 3D objects. In a traditional CAD pipeline, design engineers commonly draw a 2D wireframe\footnote{We use the terms ``wireframe'' and ``line drawing'' interchangeably.} of the desired object when creating their ideas and when communicating their ideas to others. Therefore, there is a strong need to develop algorithms which can convert 2D line drawings into 3D solid models for analysis, simulation, and manufacturing.

\begin{figure}[t]
    \centering
    \begin{tabular}{c}
        \includegraphics[height=1in]{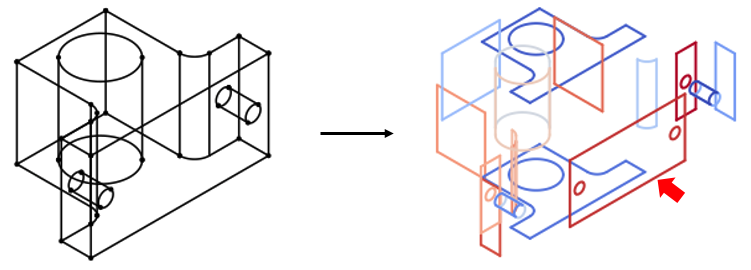} \\
        \includegraphics[height=1in]{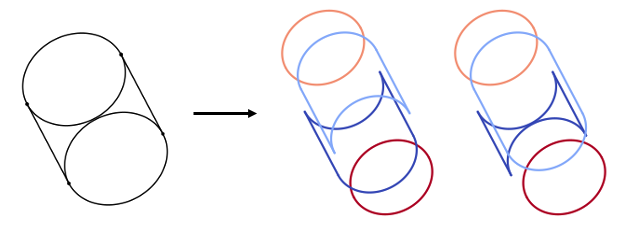}
    \end{tabular}
    \caption{Given a 2D line drawing, face identification aims to find edge loops which correspond to actual faces of the 3D object. {\bf First row:} A case where several faces (including the one indicated by a red arrow) are enclosed by two or more loops. {\bf Second row:} A case where multiple topologically correct interpretations exist.}
    \label{fig:teaser}
\end{figure}

An important step of 3D reconstruction from 2D line drawings is face identification, that is, finding loops of the edges which correspond to faces of the 3D object (Figure~\ref{fig:teaser}). If the correct face configuration of an object can be obtained, the number of degrees of freedom in its reconstruction will be greatly reduced. Many studies have been performed to achieve this goal, and in certain cases, a satisfactory solution exists. For example, it is well known that there is a unique planar embedding when the object has genus 0 and the drawing is 3-connected, from which the set of faces can be determined~\cite{ShpitalniL96}. However, for a general manifold object with complex geometry (\eg, curved surfaces) and topology (\eg, high genus), the task remains difficult.

A close look at current methods reveals two primary sources of challenges. The first one is the algorithmic complexity. As no analytical solution is available for a general manifold object, these methods need to search through the space of all possible face loops, which grows exponentially with the number of vertices. In order to make the solution efficient, much effort has been made to design heuristic search algorithms~\cite{LiuLC02, LiuT04, VarleyC10, FangLL15}. The second one is the inherent ambiguity in reconstructing 3D objects from 2D projections. For example, it is impossible for a topological algorithm to tell if two nested cycles (Figure~\ref{fig:teaser}, first row) are coplanar, thus are two loops of the same face. Besides, a single 2D projection may yield multiple topologically correct solutions (Figure~\ref{fig:teaser}, second row). In such cases, additional heuristics (\eg, number of cycles, regularities) or human intervention is required to choose the final output.

Nevertheless, humans can effortlessly ``see'' the real faces in a typical line drawing, including those in Figure~\ref{fig:teaser}. Rather than performing a tedious search for the face loops, our ability to quickly identify the faces seems to be attributed to our past experience interacting with 3D objects. This makes us wonder: Is it possible for a computer to learn to recognize faces in a data-driven fashion? This work represents a first attempt to answer the question. We train a deep neural network to detect faces using a large collection of 3D objects and their 2D projections. This way, we are able to avoid exploring a search space of exponential order. Moreover, by learning from ground truth 3D data, our method implicitly learns to generate the most plausible solutions, resolving the inherent ambiguity associated with the problem.

To this end, we cast face identification as a sequence generation problem, leveraging the natural order of co-edges in a face loop (see Section~\ref{sec:problem} for details). Given a set of co-edges in a 2D line drawing, we train a variant of the popular Transformer model~\cite{VaswaniSPUJGKP17} to predict one co-edge index that forms the face at one timestamp. For each detected face, we further classify it into different types such as planes and cylindrical surfaces. On the public ABC dataset~\cite{KochMJWABAZP19}, our model achieves $93.8\%$ and $95.9\%$ in precision and recall, respectively. Finally, with the detected faces, we develop a simple convex optimization scheme to reconstruct structured 3D models from a single 2D line drawing. 

In summary, the {\bf main contributions} of this work are two-fold: (i) We propose a first data-driven approach to face identification and study its advantages and disadvantages over existing geometry- and topology-based methods. (ii) We develop a simple scheme to reconstruct a 3D model from a single 2D line drawing using the face identification results. With this work, we discuss new opportunities for incorporating learning-based approaches into established CAD pipelines, such as identifying conflicting face loops in a geometric constraint system for 3D modeling.

\section{Related Work}
\label{sec:related}

\noindent{\bf Face identification and 3D reconstruction.} Face identification is a long-standing problem in the area of automatic interpretation of line drawings. Research in this field dates back to the '70s and '80s. Early work~\cite{MarkowskyW80} proposes a geometric approach to detecting planar faces in a 3D wireframe: it first generates possible planes at a vertex joined by two non-collinear edges, then searches for other vertices lying on each such plane and tries to use the vertices to form cycles. Subsequent work~\cite{AbbasinejadJA11, ZhuangZCJ13} can deal with more general curve networks with highly curved faces and complex topology (\ie, manifolds with high genus). But these methods rely on 3D coordinates of the wireframe, therefore cannot be applied to 2D projections.

Another line of work addresses the problem from a graph-topological point of view. To find the faces of a vertex-edge graph, \citet{Hanrahan82, DuttonB83} propose to compute the planar embedding of the graph, where the resulting regions represent the faces.
\citet{Ganter83, BrewerC86} generate an initial cycle basis from the spanning tree of the graph, then perform a cycle reduction procedure to find the faces. But these methods are only suitable for genus-0 manifolds. For manifold objects of a non-zero genus or non-manifold objects, \citet{ShpitalniL96} present a method which searches through the set of all possible sets of face loops and use various geometric criteria to assess them. The method could be slow as the number of possible face loops is $O(e^v)$ and any topological algorithm for finding faces of objects of genus $>0$ has exponential complexity~\cite{BagaliW95}. To reduce the search complexity, \citet{LiuLC02} propose a depth-first search that is primarily guided by topological constraints; \citet{LiuT04} develop a genetic algorithm which uses 2D geometric information in its assessment criteria; \citet{VarleyC10} use a heuristic search based on the shortest-path and Dijkstra's algorithm, and \citet{FangLL15} introduce a fast algorithm via edge decomposition.

Given the face topology, several studies~\cite{ShpitalniL96b, LiuCLT08, WangCLT09} reconstruct the 3D object from a single 2D line drawing. These works solve for the 3D shape in an optimization framework, using various constraints based on structural regularities, such as minimum standard deviation of angles (MSDA), face planarity, line parallelism, and corner orthogonality. 
In this work, we introduce a different method which uses the predicted face types -- information previous methods do not have access to. Further, we study the impact of imperfect face detection results on 3D reconstruction.

\smallskip
\noindent{\bf Deep models.} Our technical approach is inspired by the advance in sequence-to-sequence modeling~\cite{BahdanauCB15, VaswaniSPUJGKP17}, which has produced the state-of-the-art results in a wide range of NLP and vision tasks lately.
Specifically, our network design follows Pointer Net~\cite{VinyalsFJ15}, which proposes an autoregressive model to generate a distribution on a given input data set. PolyGen~\cite{NashGEB20} extends this idea to generate 3D polygon meshes by sequentially predicting the vertices and faces using a Transformer-based architecture.

Several recent papers also apply deep networks to CAD data~\cite{JayaramanSLWDSM21, LambourneWJSMS21, WillisPLCDLSM21, WuXZ21, WillisJLCP21, ParaBGKMGW21}. While our method shares common aspects with these work, such as network design~\cite{WuXZ21, WillisJLCP21, ParaBGKMGW21} and the use of co-edges~\cite{LambourneWJSMS21}, we tackle a different problem in face identification in this paper.

\section{Problem Formulation}
\label{sec:problem}

In this paper, a 2D wireframe projection is assumed to be an orthographic projection where all the edges (including silhouettes) and vertices of the object are visible. We make a few assumptions about the input line drawing: 
\emph{First}, the hidden lines and vertices are given. \emph{Second}, the crossing point of two edges in a line drawing is not a vertex and cannot be used to form faces. As such, the input line drawing can be represented as an edge-vertex graph $\G = (V, E)$, where each edge (or vertex) of the graph corresponds to exactly one edge (or vertex) of the object. Note that the graph may contain one or more connected components. In practice, these graphs may be a result of previous processing of a rough sketch or a scanned-in drawing~\cite{ShpitalniL97}.

\begin{figure}[t]
    \centering
    \includegraphics[width=2.0in]{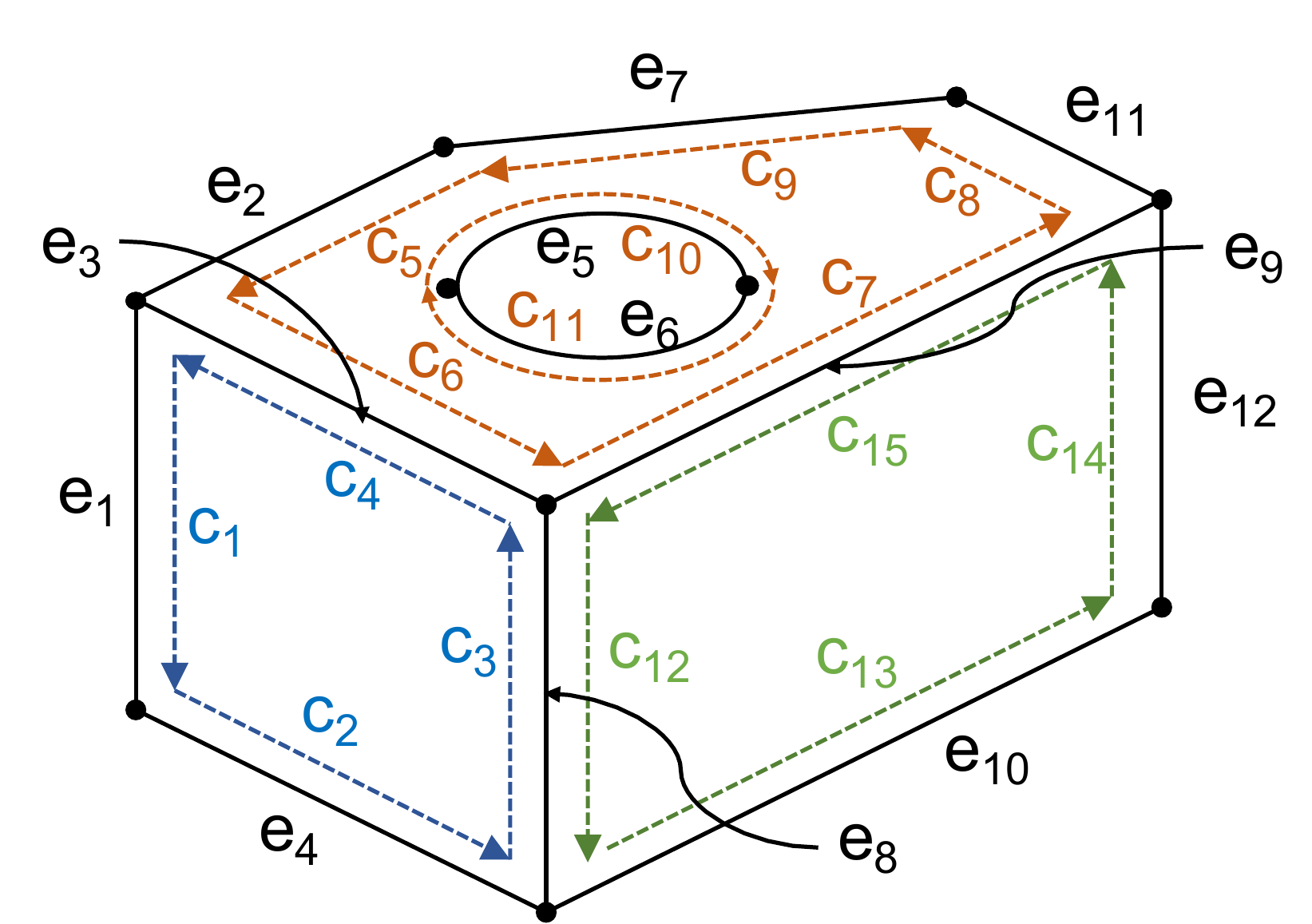}
    \caption{B-rep of an object. Hidden lines are omitted.}
    \label{fig:brep}
\end{figure}

Given the graph $\G = (V, E)$, the goal of face identification is to find all faces $F = \{f_1, \ldots, f_M\}$ of the object, where each face can be written as the set of enclosing edges: $f_i = \{e_{i_1}, \ldots, e_{i_n}\}$.
Same as prior work~\cite{LiuLC02, VarleyC10}, we focus on \emph{manifold objects} only. A manifold object is defined as a solid where every point on its surface has a neighborhood topologically equivalent to an open disk in the 2D Euclidean space. A key property of manifold objects is that each edge of a manifold is shared exactly by two faces. This property is best expressed in terms of \emph{co-edges}, $C = \{c_1, c_2, \ldots\}$, an important type of topological entities in the B-rep. As illustrated in Figure~\ref{fig:brep}, there is a co-edge pointing from vertex $v_i$ to $v_j$ if and only if there is an edge connecting $v_i$ and $v_j$. For example, $c_3$ and $c_{12}$ are two mutually mating co-edges associated with edge $e_8$.

A face can be conveniently represented as a loop (\ie, closed path) of co-edges. For example, in Figure~\ref{fig:brep}, the loops $(c_1, c_2, c_3, c_4)$ and $(c_5, c_6, c_7, c_8, c_9, c_{10}, c_{11})$ represent two faces of the object. We follow the conventional definition of loop direction: if a loop is viewed along its direction, with the face normal pointing upwards, then the face that owns the loop is to the left.

\section{Face Identification via Sequence Generation}

\begin{figure}[t]
    \centering
    \includegraphics[width=0.99\linewidth]{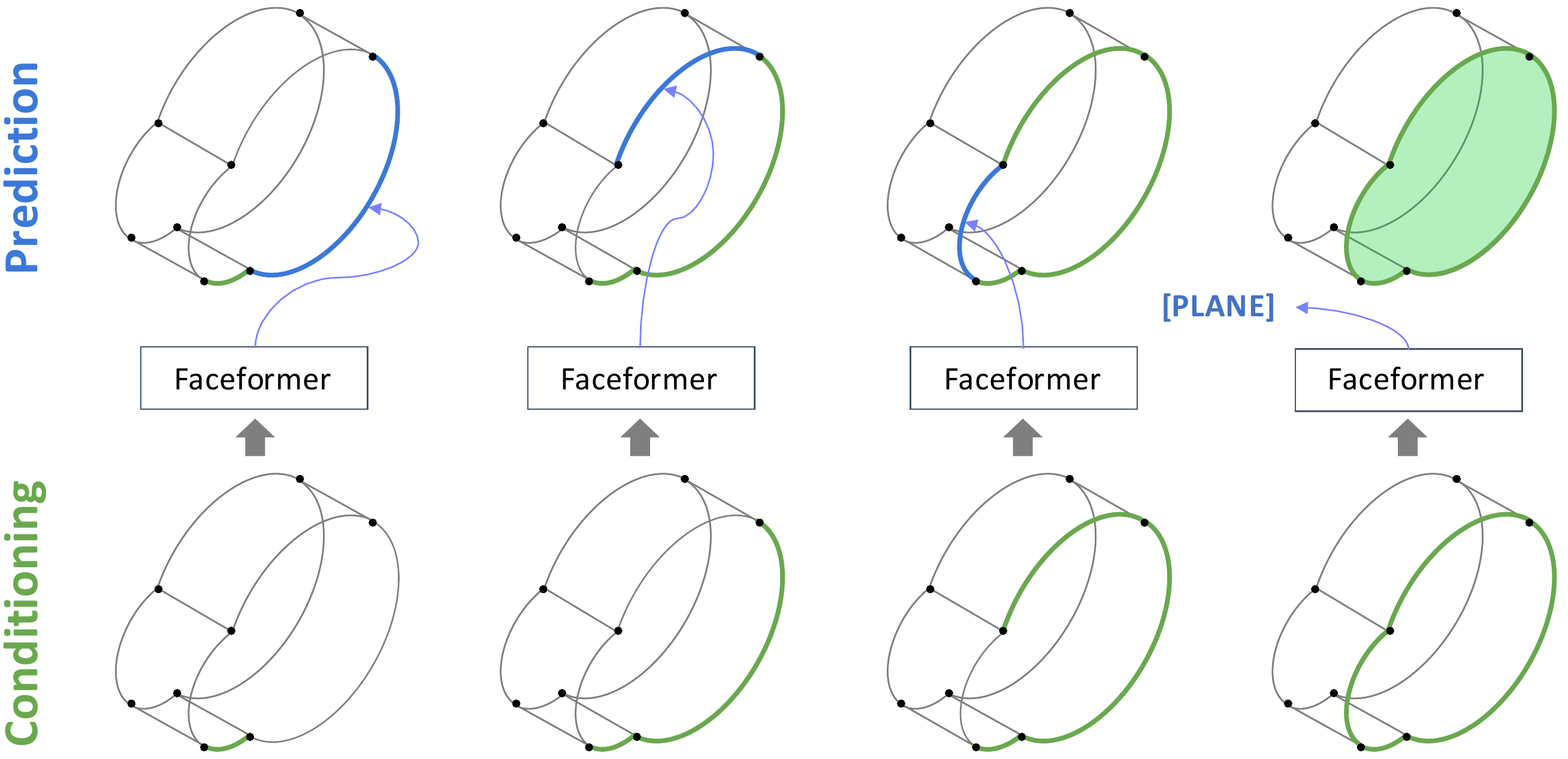}
    \caption{Our model, Faceformer, takes as input the set of all co-edges, and current sequence of co-edge indices, and outputs a distribution over the co-edge indices.}
    \label{fig:pipeline}
\end{figure}

The natural order of co-edges corresponding to a face motivates us to treat face identification as a sequence generation problem. Specifically, a face with $n$ co-edges can be written as: $f_i = (c_{i_1}, \ldots, c_{i_n})$, where each index $i_t$, $t=1, 2, \ldots, n$, is an integer between 1 and $N$, and $N$ is the total number of co-edges. Thus, starting with an arbitrary co-edge $c_{i_1}$, our goal is to grow it into a sequence of co-edges $(c_{i_1}, \ldots, c_{i_n})$ on which $c_{i_1}$ lies, as shown in Figure~\ref{fig:pipeline}. To detect all the faces $F = \{f_1, \ldots, f_M\}$, we may use every co-edge in $C$ as the starting co-edge and repeat the process for $N$ times.

In the following, we describe how to generate a single face $f_i$ given the starting co-edge $c_{i_1}$.

\subsection{Face Identification Model}

Since our goal is to select a subset of the input co-edges to form the output face, we first briefly review Pointer Net~\cite{VinyalsFJ15}, a sequence-to-sequence (seq2seq) model which uses the attention mechanism to create pointers to input elements. Given the input $P$, the main idea is to learn the conditional probability $p(\I \mid P)$ using a parametric model, such as LSTM~\cite{HochreiterS1997} or Transformer~\cite{VaswaniSPUJGKP17}, to estimate the terms of the probability chain rule:
$p(\I \mid P) = \prod_{t=1}^T p (i_t \mid i_1, \ldots, i_{t-1}, P)$.
Here, $P = \{\p_1, \p_2, \ldots\}$ is a sequence of vectors and $\I = (i_1, \ldots, i_T)$ is a sequence of $T$ indices, each between $1$ and $|P|$.

Similar to most seq2seq models, Pointer Net adopts an encoder-decoder architecture. First, it obtains a contextual embedding $\w_k$ for each input using an encoder. At each decoding time step $t$, the decoder outputs a pointer vector $\u_t$, which is then compared to the contextual embeddings via a dot-product. The resulting scores are normalized using a softmax to form a valid distribution over the input set:
\begin{align}
    \{\w_k\}_{k=1}^{N} & = \mathbf{Encoder}(P; \theta) \\
    \u_t & = \mathbf{Decoder}(\I_{<t}, P; \theta) \\
    p(i_t = k \mid \I_{<t}, P; \theta) & = \mathbf{softmax}_{k}(\u_t^T \w_k)
\end{align}
Finally, the parameters of the model are learned by maximizing the conditional probabilities on a training set.

\smallskip
\noindent\textbf{Input sequence and embeddings.} In our problem, the input sequence consists of all co-edges $C$. To further leverage geometric cues, we propose to classify the faces into different types -- a benefit of data-driven approaches. In this work we only consider two special face types, namely planar surface and cylinder surface, but the method can be easily extended to other types. To this end, we add three special tokens, \texttt{[PLANE]}, \texttt{[CYLINDER]} and \texttt{[OTHERS]}, to indicate different face types. We replace regular stop tokens with the face type tokens, expecting the network to predict one of the three tokens immediately after generating all co-edges of the face.

We jointly embed the special tokens with the co-edges, to obtain a total of $N+3$ input embeddings. We use two embeddings for each input co-edge, including (i) value embedding, representing the coordinate value of the edge, and (ii) position embedding, indicating the token location in the sequence. Due to the varying edge length, we uniformly sample a fixed number of edge points to represent the co-edge as in~\cite{DasYHXS20}. The points are ordered based on the co-edge direction.
Then, we flatten the edge points and apply two linear layers to obtain a 512-dimensional embedding.

\smallskip
\noindent\textbf{Output sequence and embeddings.} As mentioned above, we represent each face as a sequence of co-edges $f_i = (c_{i_1}, \ldots, c_{i_n})$. A face type token is added (i) to predict the face type and (ii) to indicate the end of the sequence. When a face consists of multiple loops, special care needs to be taken to ensure a unique co-edge sequence. In such cases, except for the loop to which the starting co-edge belongs, we order the co-edges in each other loop from lowest to highest first by its $x$-coordinate, followed by $y$-coordinate. Take Figure~\ref{fig:brep} as an example, if we use $c_{11}$ as the starting co-edge, the desired face sequence should be $(c_{11}, c_{10}, c_5, c_6, c_7, c_8, c_9)$.

Similar to input embeddings of the encoder, we use learned position and value embeddings for the inputs to the decoder. We use the co-edge's contextual embedding from the encoder output as its value embedding.

\smallskip
\noindent\textbf{Network architecture.} Inspired by recent success of Transformer-based architectures~\cite{VaswaniSPUJGKP17}, we adopt it as the basic block of our face identification model. Given the input co-edge embeddings, the encoder encodes them into contextual embeddings, then the decoder outputs pointers based on the contextual embeddings and the decoder input. The model consists of 6 Transformer encoder and decoder layers, with a feed-forward dimension of 1024 and 8 attention heads.

\subsection{Training and Inference}
\label{sec:method:implementation}

We implement our model with PyTorch and PyTorch Lightning. We use Adam optimizer~\cite{KingmaB15} with a learning rate of \num{1e-4}. The batch size is set to \num{4}. The model is trained with \num{400000} iterations, taking about \num{30} hours to converge on a single NVIDIA RTX 3090 GPU device.

At inference time, since our parallel model predicts faces from all co-edges independently, there are many duplicated faces in the result. We take three post-processing steps to remove invalid and duplicated predictions. First, because a face is defined as a closed path of co-edges, we filter out any unclosed face predictions. Second, we remove predictions that contain both co-edges associated with the same edge, as this never happens in real faces. Finally, we identify duplicated predictions by comparing the set of edges generated in each face prediction. Note that these duplicated predictions may have different face type classifications. We count the number of times each face type is predicted, and take the face type of the highest count as its predicted face type.

\subsection{Experiments}

\subsubsection{Experimental Setup}

\noindent \textbf{Dataset.} We build a benchmark for this novel task using a subset of CAD mechanical models from ABC dataset~\cite{KochMJWABAZP19}. We use pythonOCC\footnote{\url{https://github.com/tpaviot/pythonocc}}, a Python 3D development framework built upon the Open CASCADE Technology, to project CAD models into 2D line drawings. We first normalize the shape such that the half  diagonal length of the bounding box is equal to 1. Then, the camera is placed at a random distance between \num{1.25} to \num{1.5} away from the object center, pointing towards the object. The viewpoints are randomly sampled on a hemisphere. The dataset consists of \num{9370}/\num{202}/\num{504} samples for training/validation/testing.

To eliminate cases that are unnecessarily complicated, we filter out shapes with more than \num{42} faces or \num{37} edges in a face. Since the ABC dataset has many duplicate shapes, we also run additional filters based on the shape's topology and three orthogonal views to remove duplicates.

\smallskip
\noindent \textbf{Evaluation metrics.} To evaluate the performance of face identification, we compute the \emph{precision} and \emph{recall} at the face level. We treat each face as an unordered set of edges. For our model, this is the result after the post-processing step as described in Section~\ref{sec:method:implementation}. A prediction $f$ is said to match a ground truth face $f^*$ if and only if the two sets are equal. Then, let $F^*$ denote the set of ground truth faces, and $F$ denote the set of faces detected by any method, the precision and recall are defined as:
$\textup{precision} = |F \bigcap F^*|/|F|, \textup{recall} = |F \bigcap F^*|/|F^*|$.

For face type classification, we report the \emph{classification rate}, which is the percentage of correctly classified faces among all faces correctly detected by our model. 

\subsubsection{Experimental Results}

\noindent \textbf{Comparison with prior work.} We first compare our method with an existing method, FindingFaces~\cite{VarleyC10}. To the best of our knowledge, this is the only related work with public code or implementation\footnote{\url{http://www.regeo.uji.es/FindingFaces.htm}}. It assumes that, for each co-edge, the true face has the least cost (\eg, the shortest path) among all loops enclosing that co-edge. Then, the Dijkstra's algorithm can be applied to find the least-cost closed loops. But in practice, not all faces correspond to the shortest paths, so a more complicated ``cost'' is proposed in~\cite{VarleyC10}. Besides, the order in which the co-edges are considered also affects its performance. Thus, much effort is made to develop heuristic rules for edge priorities.

\begin{table}[t]
    \centering
    \begin{tabular}{l|ccc}
        \toprule
        method & precision & recall & runtime (s) \tabularnewline
        \midrule
        FindingFaces~\cite{VarleyC10} & 97.3 & \textbf{97.8} & \textbf{0.008} \tabularnewline
        Ours & \textbf{98.2} & 97.7 & 0.119 \tabularnewline
        \bottomrule
    \end{tabular}
    \caption{Comparison with prior work (on the subset of polyhedral objects only). FindingFaces does not work on non-polyhedra.}
    \label{tab:result:comp}
\end{table}

Despite its efficiency, a key limitation of FindingFaces~\cite{VarleyC10} is that it is applicable to polyhedral objects only. Among the 504 randomly sampled objects in our test set, only 126 ($25\%$) are polyhedra. As shown in Table~\ref{tab:result:comp}, both methods achieve high accuracies on this subset. Comparing to results on the full test set in Table~\ref{tab:result:arch}, one can infer that polyhedra are relatively simple cases. Among the mistakes made by FindingFaces, a notable issue is that it cannot detect faces with more than one loop (\ie, objects with holes) -- a common problem for all topological approaches.

\begin{table}[t]
    \centering
    \begin{tabular}{l|ccc}
        \toprule
        models & precision & recall & runtime (s) \tabularnewline
        \midrule
        Seq2seq & 77.6 & 76.8 & 0.69 \tabularnewline
        Seq2seq + co-edge & \textbf{94.9} & 90.5 & 0.65\tabularnewline
        Ours & 93.8 & \textbf{95.9} & \textbf{0.15}\tabularnewline
        \bottomrule
    \end{tabular}
    \caption{Experiment results on network design.}
    \label{tab:result:arch}
\end{table}

\smallskip
\noindent \textbf{Experiment on network design.} In this experiment, we compare our model with two variants of it to illustrate the benefits of (i) using co-edges and (ii) face-wise parallel prediction.

The first variant, \emph{seq2seq}, directly operates on the edges (instead of co-edges) and generates all faces in a sequential fashion. In this variant, the input sequence would be the set of all edges, ordered from lowest to highest first by its $x$-coordinate, then by $y$-coordinate. For the output sequence, we introduce three additional special tokens, \texttt{[SOS]}, \texttt{[EOS]}, and \texttt{[SEP]}, to indicate the start and end of sequence, and the separation between faces, respectively. During training, we order faces according to the edge indices and then concatenate all faces into a single sequence. Take Figure~\ref{fig:brep} as an example, the desired output sequence would be $({\tt SOS}, e_1, e_3, e_4, e_8, {\tt SEP}, e_2, e_3, e_5, e_6, e_7, e_9, e_{11}, {\tt SEP}, e_8,\\ e_9, e_{10}, e_{12}, {\tt EOS})$.

The second variant, \emph{seq2seq + co-edge}, also generates all faces sequentially but uses co-edges. In this variant, the input sequence would include all the co-edges and three special tokens \texttt{[SOS]}, \texttt{[EOS]}, and \texttt{[SEP]}. We order the faces in the same way as the first variant. Thus, the desired output sequence for the example in Figure~\ref{fig:brep} is $({\tt SOS}, c_1, c_2, c_3, c_4, {\tt SEP}, c_5, c_6, c_7, c_8, c_9, c_{10}, c_{11}, {\tt SEP}, c_{12},\\ c_{13}, c_{14}, c_{15}, {\tt EOS})$.

Table~\ref{tab:result:arch} shows quantitative results on the full test set. The seq2seq model using edges performs substantially worse because (i) unlike co-edges, edges do not encode directional information, so the model cannot take advantage of the natural order of face loops; and (ii) an edge is shared by two faces, so there is ambiguity about which face to predict given an edge. Comparing the two models using co-edges, the one with parallel prediction has slightly lower precision ($-1.1\%$) but higher recall ($+5.4\%$). This is because it makes a prediction starting from each co-edge, thus would cover each face several times. In contrast, the model with sequential prediction only generates each face once. We also record the average runtime of the models for one shape in the test set. As one can see in Table~\ref{tab:result:arch}, the model with parallel prediction is much faster.

Note that our focus here is on the various ways to detect face loops, thus we do not perform face type classification in this experiment. To enable joint face type classification for the two variants, one can simply replace the \texttt{[SEP]} token with the corresponding face type tokens. Based on our experience, enabling face type classification has negligible impact on the face identification results.

\begin{table}[t]
    \centering
    \begin{tabular}{l|cccc}
        \toprule
        input data & precision & recall & cls. rate \tabularnewline
        \midrule
        Ours  & 93.8 & 95.9 & 97.5 \tabularnewline
        Ours - perspective  & 93.6 & 96.2 & 97.6 \tabularnewline
        Ours - fixed viewpoint & 95.9 & 97.3 & 97.6 \tabularnewline
        \bottomrule
    \end{tabular}
     \caption{Experiment results on input data.}
    \label{tab:result:data}
\end{table}

\begin{figure*}[t]
    \centering
        \includegraphics[width=0.105\linewidth]{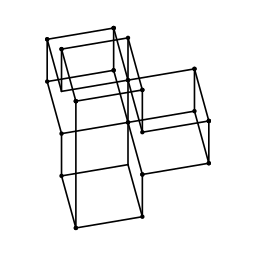}
        \includegraphics[width=0.105\linewidth]{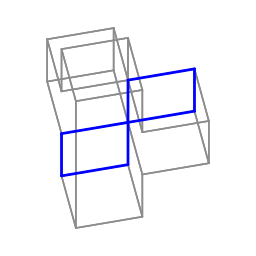} \vrule 
        \includegraphics[width=0.105\linewidth]{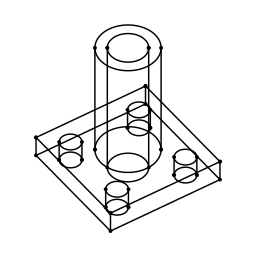}
        \includegraphics[width=0.105\linewidth]{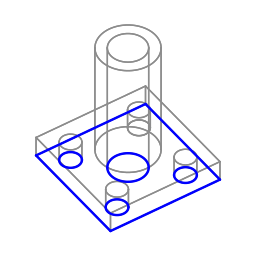}
        \includegraphics[width=0.105\linewidth]{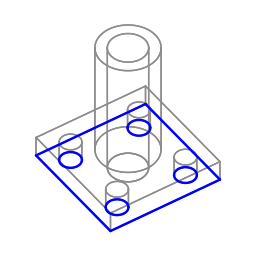} \vrule 
        \includegraphics[width=0.105\linewidth]{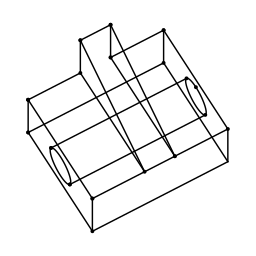}
        \includegraphics[width=0.105\linewidth]{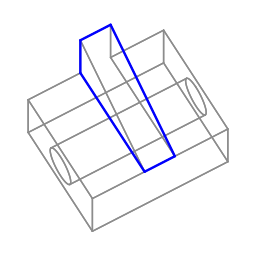} \vrule 
        \includegraphics[width=0.105\linewidth]{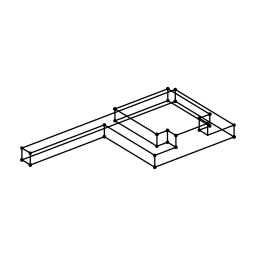}
        \includegraphics[width=0.105\linewidth]{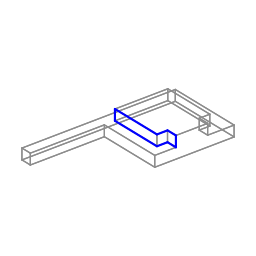}
        \includegraphics[width=0.105\linewidth]{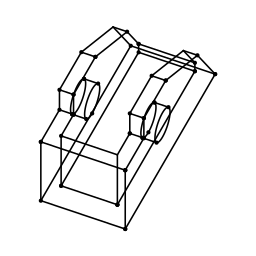}
        \includegraphics[width=0.105\linewidth]{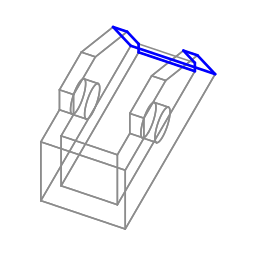}
        \includegraphics[width=0.105\linewidth]{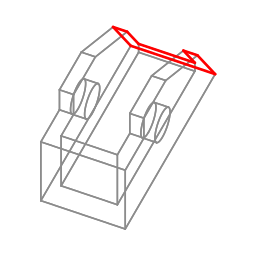} \vrule 
        \includegraphics[width=0.105\linewidth]{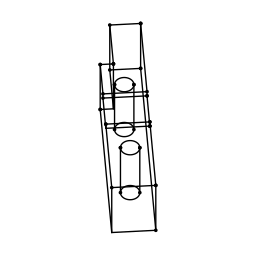}
        \includegraphics[width=0.105\linewidth]{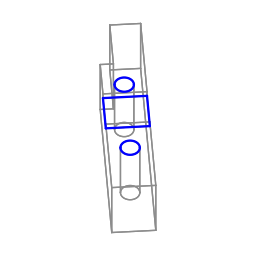} \vrule 
        \includegraphics[width=0.105\linewidth]{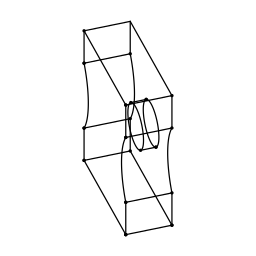}
        \includegraphics[width=0.105\linewidth]{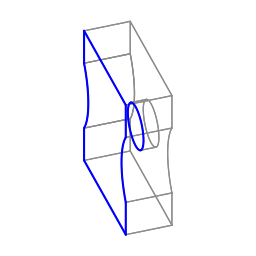}
        \includegraphics[width=0.105\linewidth]{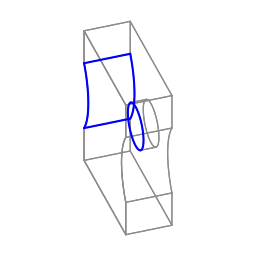}
        \includegraphics[width=0.105\linewidth]{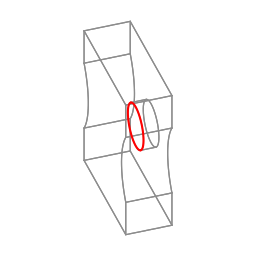}
        \includegraphics[width=0.105\linewidth]{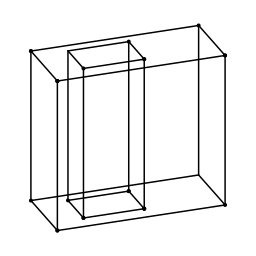}
        \includegraphics[width=0.105\linewidth]{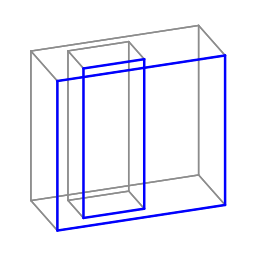}
        \includegraphics[width=0.105\linewidth]{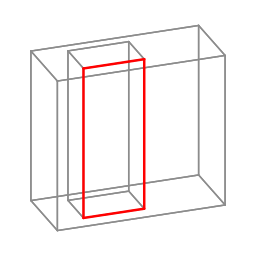} \vrule
        \includegraphics[width=0.105\linewidth]{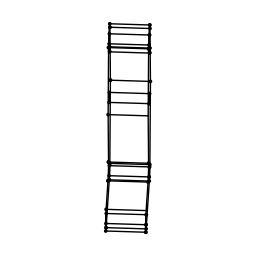}
        \includegraphics[width=0.105\linewidth]{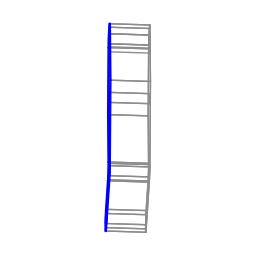}
        \includegraphics[width=0.105\linewidth]{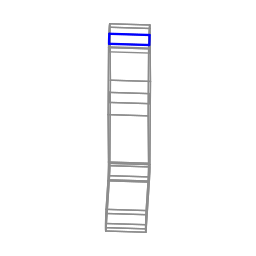}
        \includegraphics[width=0.105\linewidth]{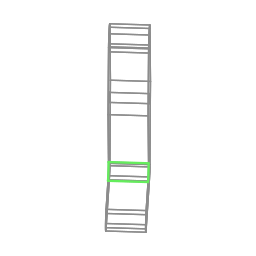}
        \includegraphics[width=0.105\linewidth]{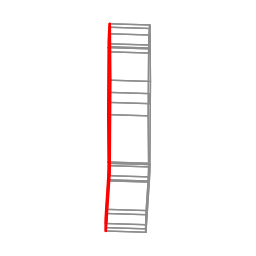}
        \includegraphics[width=0.105\linewidth]{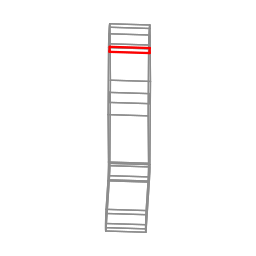}
        \includegraphics[width=0.105\linewidth]{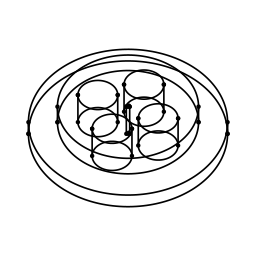}
        \includegraphics[width=0.105\linewidth]{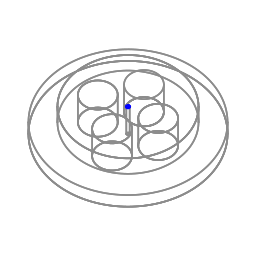}
        \includegraphics[width=0.105\linewidth]{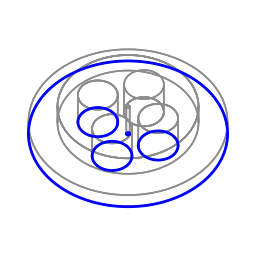}
        \includegraphics[width=0.105\linewidth]{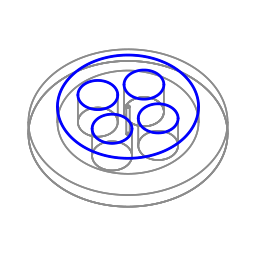}
        \includegraphics[width=0.105\linewidth]{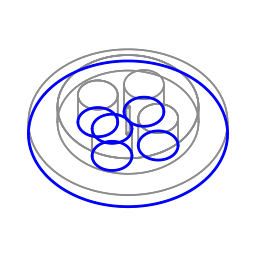}
        \includegraphics[width=0.105\linewidth]{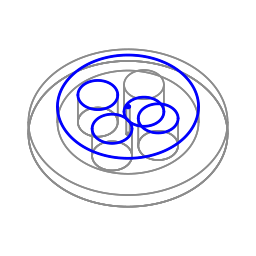}
        \includegraphics[width=0.105\linewidth]{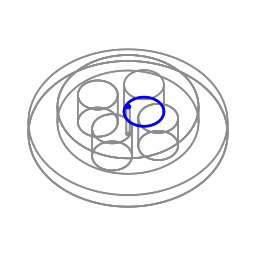}
        \includegraphics[width=0.105\linewidth]{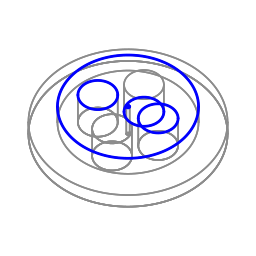}
        \includegraphics[width=0.105\linewidth]{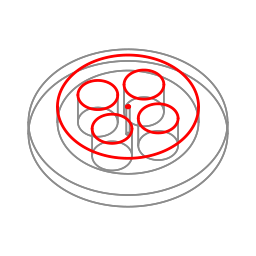}
    \caption{Incorrect predictions made by our face identification model. For each case, we show the input line drawing, followed by predictions with wrong face loops (\textcolor{blue}{blue}), predictions with wrong face types (\textcolor{green}{green}), and missed faces (\textcolor{red}{red}).}
    \vspace{-3mm}
    \label{fig:face-results}
\end{figure*}

\smallskip
\noindent \textbf{Experiment on input data.} In the next experiment, we study the performance of our model with different types of input data. First, we replace orthographic projection with perspective projection when generating the 2D line drawings. As shown in Table~\ref{tab:result:data}, this change has very small impact on all the metrics. Second, we fix the camera viewpoint to generate an isometric drawing for each shape. In CAD, an isometric view is commonly used to reveal as much information about the 3D shape as possible, and to avoid situations where the object's edges or vertices coincide (or appear as joined) accidentally. Therefore, such a viewpoint is considered easier than random viewpoints. By employing a fixed viewpoint for all objects, our model achieves even higher accuracies.

Furthermore, our model achieves high (and almost identical) classification rates with all input data types.

\smallskip
\noindent \textbf{Qualitative results.} Figure~\ref{fig:face-results} visualizes \emph{all} incorrect predictions made by our model for various objects. Some common problems are (i) incomplete prediction when a face consists of multiple nested edge loops, and (ii) grouping edges or loops which belong to different faces. But as we will see in the next section, these errors have varying impact on the reconstruction of the 3D models. 

\section{3D Object Reconstruction}
\label{sec:rec}

We now tackle the problem of 3D reconstruction with the predicted face loops and types. In the literature, this is often formulated as recovering the missing depths of the vertices of a line drawing. While knowledge about face topology significantly reduces the number of degrees of freedom, such information itself is not enough to uniquely determine the 3D geometry. Prior work~\cite{ShpitalniL96b, LiuCLT08, WangCLT09} resort to additional optimization criteria such as MSDA, face planarity, line parallelism, and corner orthogonality. We observe a few problems with these approaches: (i) the criteria are designed to emulate the human perception of a 2D line drawing as a 3D object, but it is not uncommon for them to be violated in practice; (ii) the search for optimal solutions could get stuck at local minimum; (iii) they do not use information about face types.

Our primary goal is to build 3D models from the output of a deep face identification model -- something never considered in prior work. Thus, we prefer a solution which decouples the impact of mistakes made by the network from that of other constraints or algorithms employed.

\begin{figure}[t]
    \centering
    \begin{tabular}{c|c}
    \hspace{-4mm}
        \includegraphics[height=0.95in]{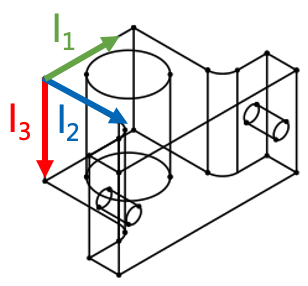} &
        \includegraphics[height=0.95in]{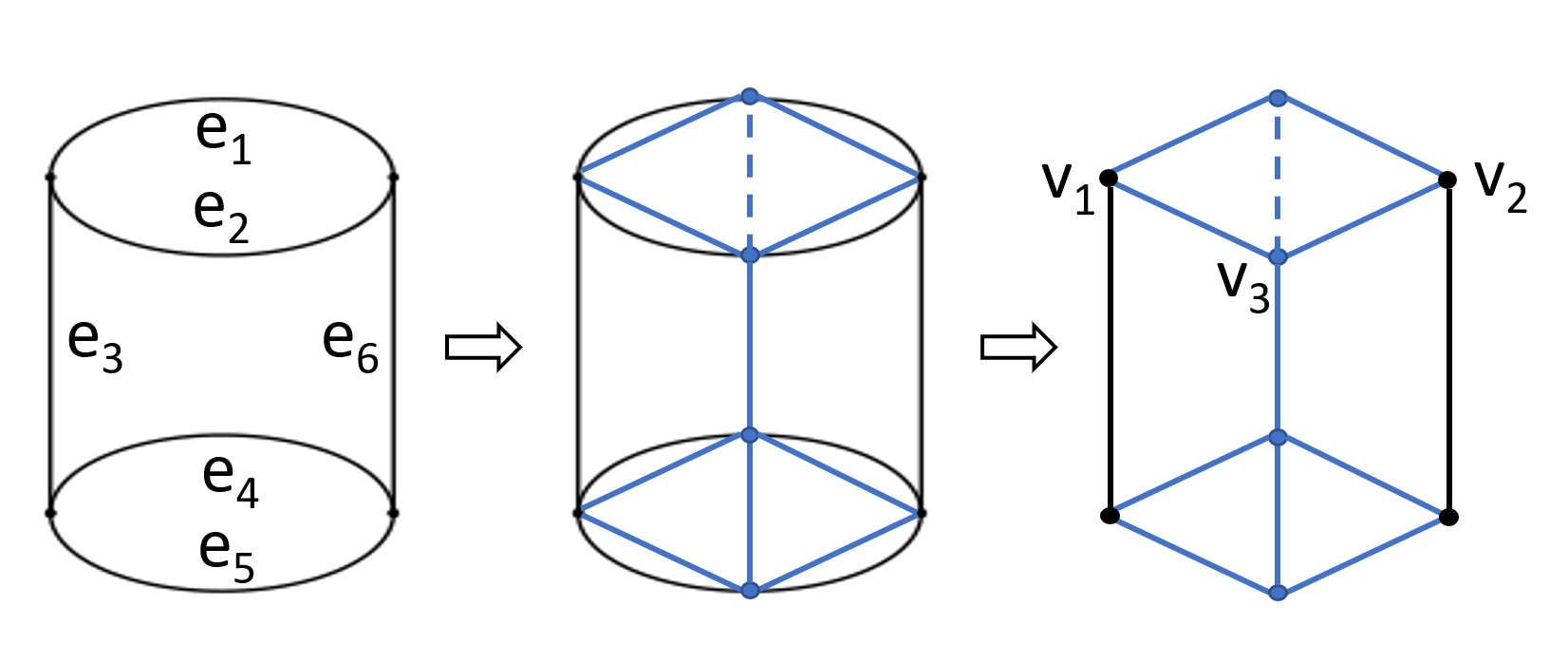} \\
        (a) & (b)
    \end{tabular}
    \caption{3D reconstruction with predicted faces. (a) Illustration of three dominant directions. (b) Handling curved surfaces.}
    \label{fig:directions}
    \vspace{-3mm}
\end{figure}

To this end, we develop a simple method that relies on only one type of constraints: line parallelism. As shown in Figure~\ref{fig:directions}(a), we assume that there are three mutually orthogonal directions $\{\l_1, \l_2, \l_3\}$ in the 3D scene, \ie, $\l_j^T \l_k =0, \forall j\neq k$. While these directions are provided \emph{a priori} in this work, techniques for automatic estimation exist: for orthographic projections, these directions can be found by grouping parallel lines in a line drawing; for perspective projections, these directions correspond to the three dominant vanishing points. Afterward, the 3D directions $\{\l_1, \l_2, \l_3\}$ can be obtained via camera calibration~\cite{HartleyR2000}.

With this assumption, we are able to align the faces to the dominant directions according to their enclosing edges and solve for the 3D geometry via convex optimization. Below we first describe our method for objects with planar faces, then extend it to curved surfaces.

\subsection{3D Reconstruction of Planar Objects}
\label{sec:rec:planar}

Suppose we are given an object with $M$ faces $F = \{f_1, \ldots, f_M\}$ and $L$ vertices $V = \{v_1, \ldots, v_L\}$. For each vertex, we write $v_l = [x_l, y_l, z_l]^T$, where $z_l$ is the unknown depth of $v_l$ in the camera coordinate system.

If the object is planar, then each face $f_i$ can be represented by the plane equation $a_i x + b_i y + z + c_i = 0$. If a vertex $v_l$ lies on two or more faces, for each pair of faces, say $(f_1, f_2)$, we have $a_1 x_l + b_1 y_l + c_1 - a_2 x_l - b_2 y_l - c_2 = 0$. For all vertices in $V$, we can write similar constraints for pairs of faces. Rewriting all these linear equations in matrix form, we have:
$\P_1 \f = 0$,
where $\f = [a_1, b_1, c_1, \ldots, a_M, b_M, c_M]^T$ is a vector consisting of all the parameters of the faces. 

For each dominant direction $\l_j, j=\{1,2,3\}$, we can identify all edges in the line drawing which are parallel to it, thus also find the faces which align with it. For such a face, say $f_1$, we have $a_1 l_x^j + b_1 l_y^j + l_z^j = 0$. Rewriting all these linear equations in matrix form, we have:
$\P_2 \f = 0$.

Combining the above constraints, we can find $\f$ by solving a convex optimization problem:
\begin{align}
\begin{aligned}
    \min_{\f} & \quad \|\P\f\|_1, \\
    \textup{s.t.} & \quad z_l = -(a_i x_l + b_i y_l + c_i) > 0, \forall \v_l \textup{ on } f_i.
    \label{eqn:3d}
\end{aligned}
\end{align}
Here, the constraints enforce that the 3D vertices lie in front of the camera.

\subsection{Handling Curved Surfaces}

To deal with curved surfaces, we follow the approach proposed in~\cite{WangCLT09}. The main idea is to replace a curve with straight line segments so that the object becomes a polyhedron. Then, methods for planar objects, such as the one described in Section~\ref{sec:rec:planar}, can be applied. Finally, the curved surface is recovered by fitting general B{\'e}zier curves to the corresponding 3D vertices.

In our case, because the face types are known, the original approach can be substantially simplified, as we (i) do not need to employ an algorithm to distinguish between curved and planar faces; (ii) can develop face approximation and fitting methods for each specific face type. 

Figure~\ref{fig:directions}(b) shows an example of converting a curved surface on a cylinder into planar faces. For face $(e_2, e_3, e_5, e_6)$, we find the singular points on curves $e_2$ and $e_5$ and then replace each curve with two straight lines. Once the 3D geometry of the polyhedron is reconstructed, the original curves can be recovered by fitting a 3D circle to the three vertices (\eg, $v_1$, $v_2$, and $v_3$ in the figure).

\subsection{Experiments}

\begin{figure}[t]
    \centering
    \begin{tabular}{c|c}
     	\toprule
        Input & Reconstructed shape \tabularnewline
        \midrule
        \includegraphics[width=0.20\linewidth]{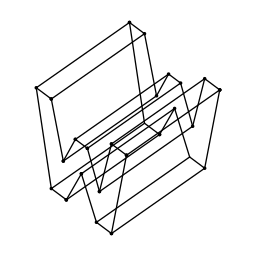} & 
        \includegraphics[width=0.20\linewidth]{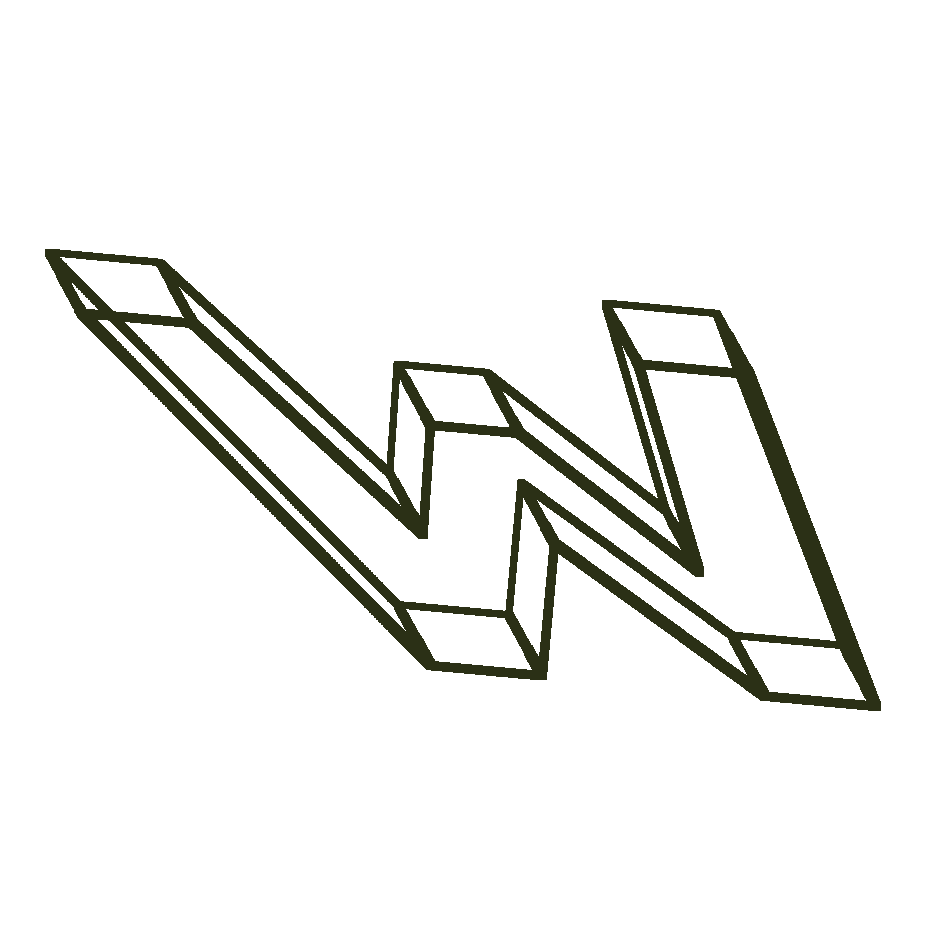} 
        \includegraphics[width=0.20\linewidth]{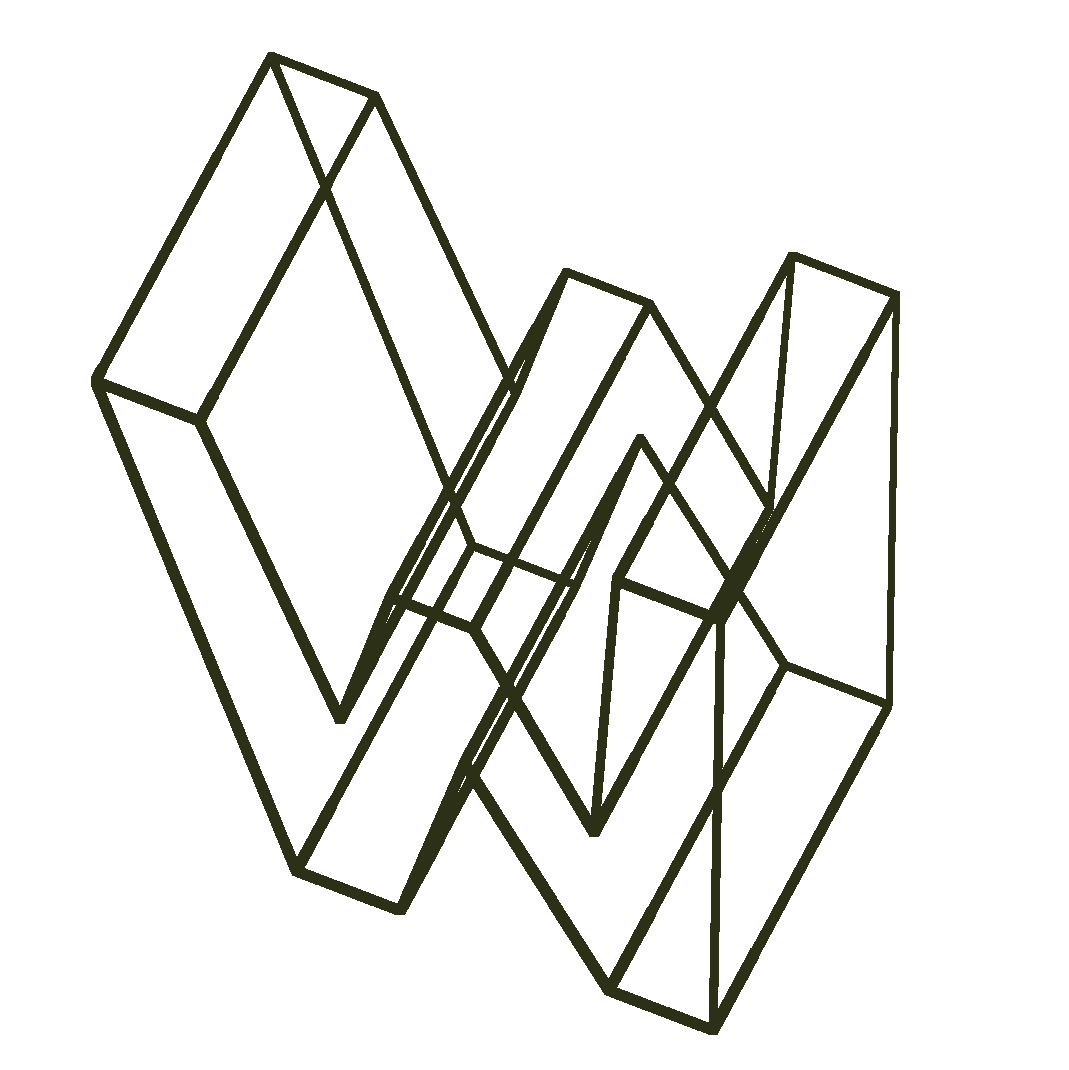} \tabularnewline
        \includegraphics[width=0.20\linewidth]{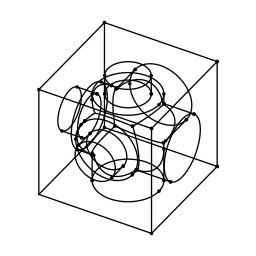} &
        \includegraphics[width=0.20\linewidth]{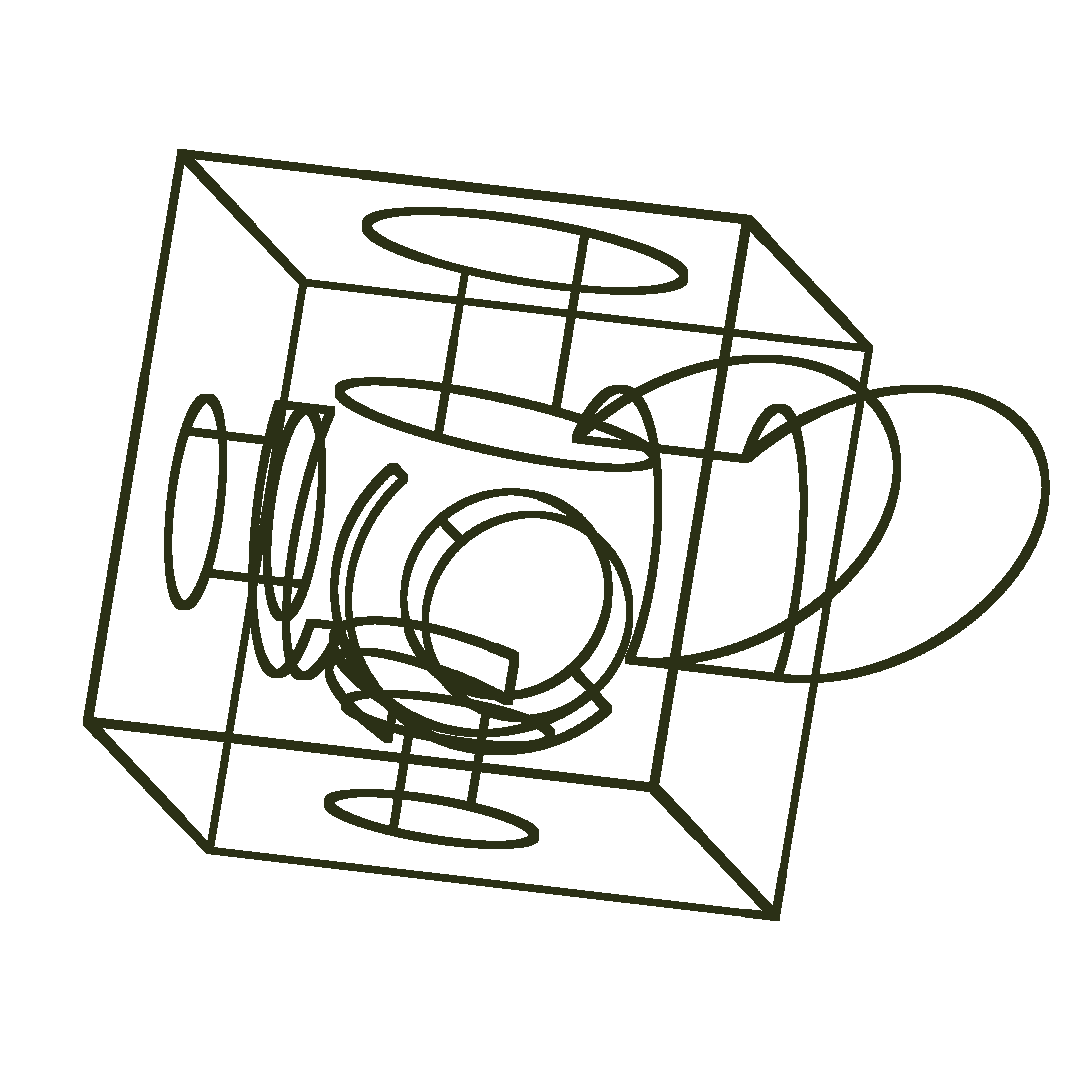} 
        \includegraphics[width=0.20\linewidth]{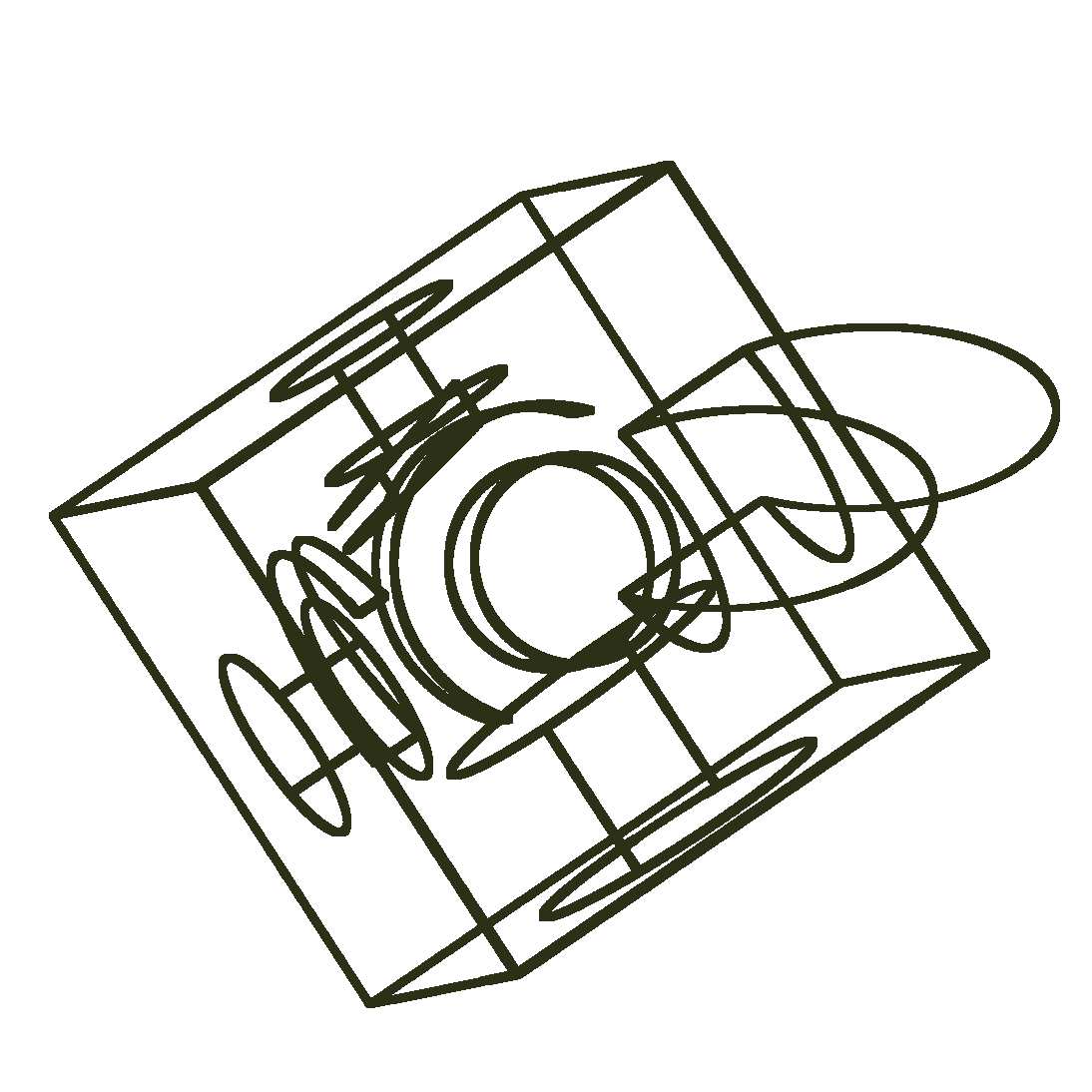} \tabularnewline
        \bottomrule
    \end{tabular}
    \caption{Two problematic cases of 3D reconstruction with ground truth faces.}
    \vspace{-3mm}
    \label{fig:failure}
\end{figure}

Before presenting the 3D reconstruction results based on the predicted faces, we point out that in rare cases ($<5\%$) the reconstruction may not be perfect even with ground truth faces, due to the simple assumption (\ie, line parallelism) we employ in the pipeline. Figure~\ref{fig:failure} shows two examples where major parts of the model do not align with any dominant direction, making the reconstruction under-constrained. As a result, the 3D model may appear to be distorted (first row) or partly incorrect (second row). In practice, these may be addressed by introducing additional constraints. But we choose to keep the reconstruction method simple so that the impact of errors in face identification can be better analyzed.

\begin{figure*}[t]
    \centering
    \begin{tabular}{c|c|c||c|c|c}
        \toprule
        Input & Ours & AtlasNet & Input & Ours & AtlasNet \tabularnewline
        \midrule
        \includegraphics[width=0.082\linewidth]{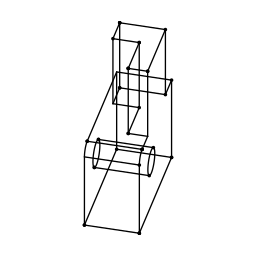} & 
        \includegraphics[width=0.082\linewidth]{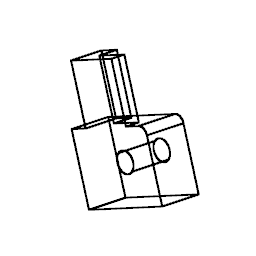} 
        \includegraphics[width=0.082\linewidth]{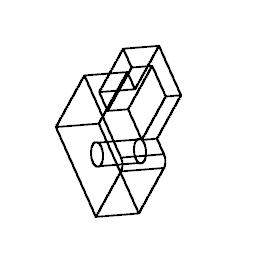} 
        \includegraphics[width=0.082\linewidth]{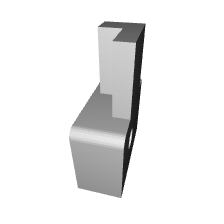} &
        \includegraphics[width=0.082\linewidth]{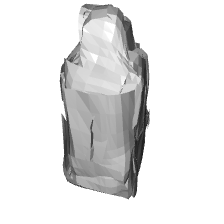} &

        \includegraphics[width=0.082\linewidth]{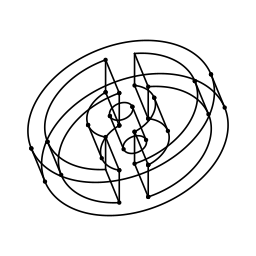} & 
        \includegraphics[width=0.082\linewidth]{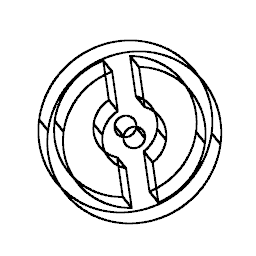} 
        \includegraphics[width=0.082\linewidth]{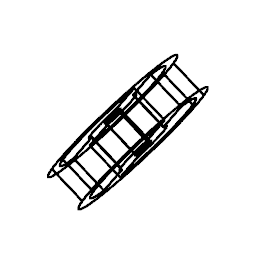}  
        \includegraphics[width=0.082\linewidth]{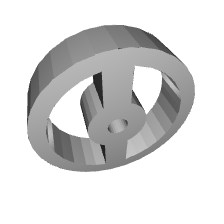} &
        \includegraphics[width=0.082\linewidth]{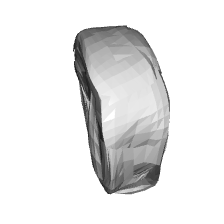} \tabularnewline

        \includegraphics[width=0.082\linewidth]{figures/reconstruction/input/00035116.png} & 
        \includegraphics[width=0.082\linewidth]{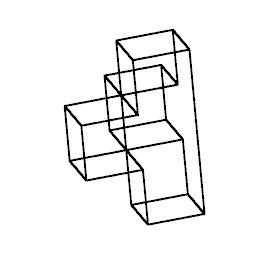} 
        \includegraphics[width=0.082\linewidth]{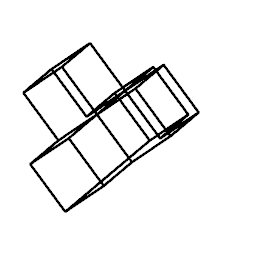}  
        \includegraphics[width=0.082\linewidth]{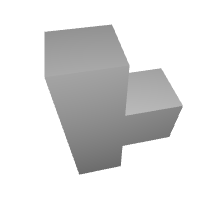} &
        \includegraphics[width=0.082\linewidth]{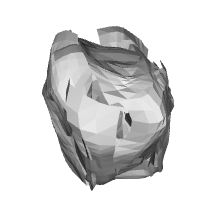} &

        \includegraphics[width=0.082\linewidth]{figures/reconstruction/input/00039626.png} &  
        \includegraphics[width=0.082\linewidth]{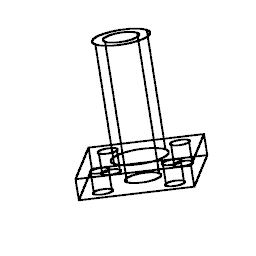} 
        \includegraphics[width=0.082\linewidth]{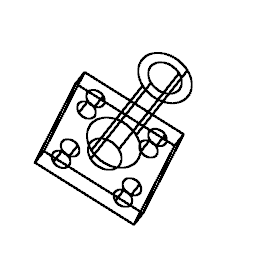}  
        \includegraphics[width=0.082\linewidth]{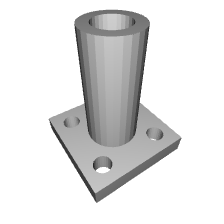} &
        \includegraphics[width=0.082\linewidth]{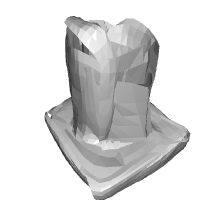} \tabularnewline

        \includegraphics[width=0.082\linewidth]{figures/reconstruction/input/00024807.png} & 
        \includegraphics[width=0.082\linewidth]{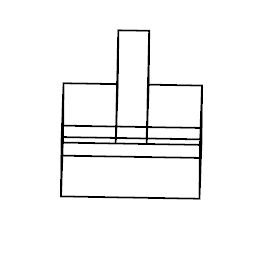} 
        \includegraphics[width=0.082\linewidth]{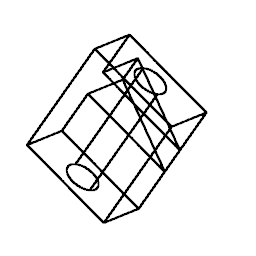}  
        \includegraphics[width=0.082\linewidth]{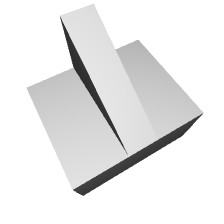} &
        \includegraphics[width=0.082\linewidth]{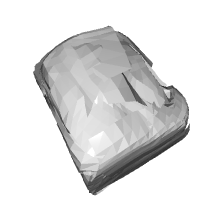} &

        \includegraphics[width=0.082\linewidth]{figures/reconstruction/input/00071288.png} & 
        \includegraphics[width=0.082\linewidth]{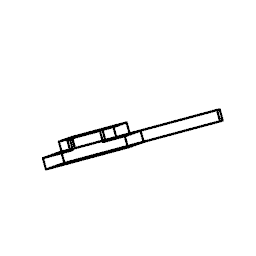} 
        \includegraphics[width=0.082\linewidth]{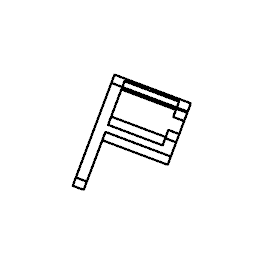}  
        \includegraphics[width=0.082\linewidth]{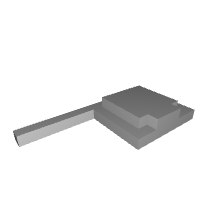} &
        \includegraphics[width=0.082\linewidth]{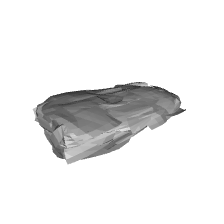} \tabularnewline
        
        \includegraphics[width=0.082\linewidth]{figures/reconstruction/input/00051162.png} &
        \includegraphics[width=0.082\linewidth]{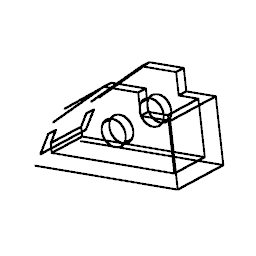} 
        \includegraphics[width=0.082\linewidth]{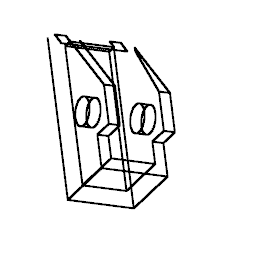}  
        \includegraphics[width=0.082\linewidth]{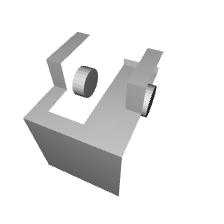} &
        \includegraphics[width=0.082\linewidth]{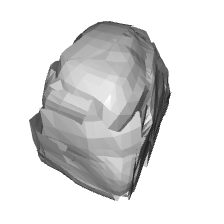} &

        \includegraphics[width=0.082\linewidth]{figures/reconstruction/input/00059834.png} &
        \includegraphics[width=0.082\linewidth]{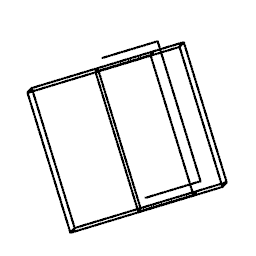} 
        \includegraphics[width=0.082\linewidth]{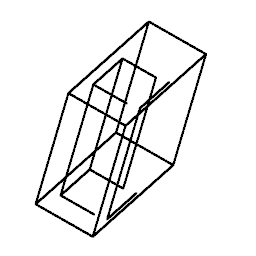}  
        \includegraphics[width=0.082\linewidth]{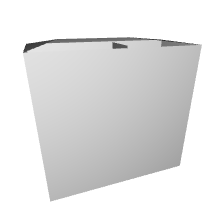} &
        \includegraphics[width=0.082\linewidth]{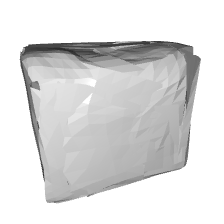} \tabularnewline

        \includegraphics[width=0.082\linewidth]{figures/reconstruction/input/00052043.png} &
        \includegraphics[width=0.082\linewidth]{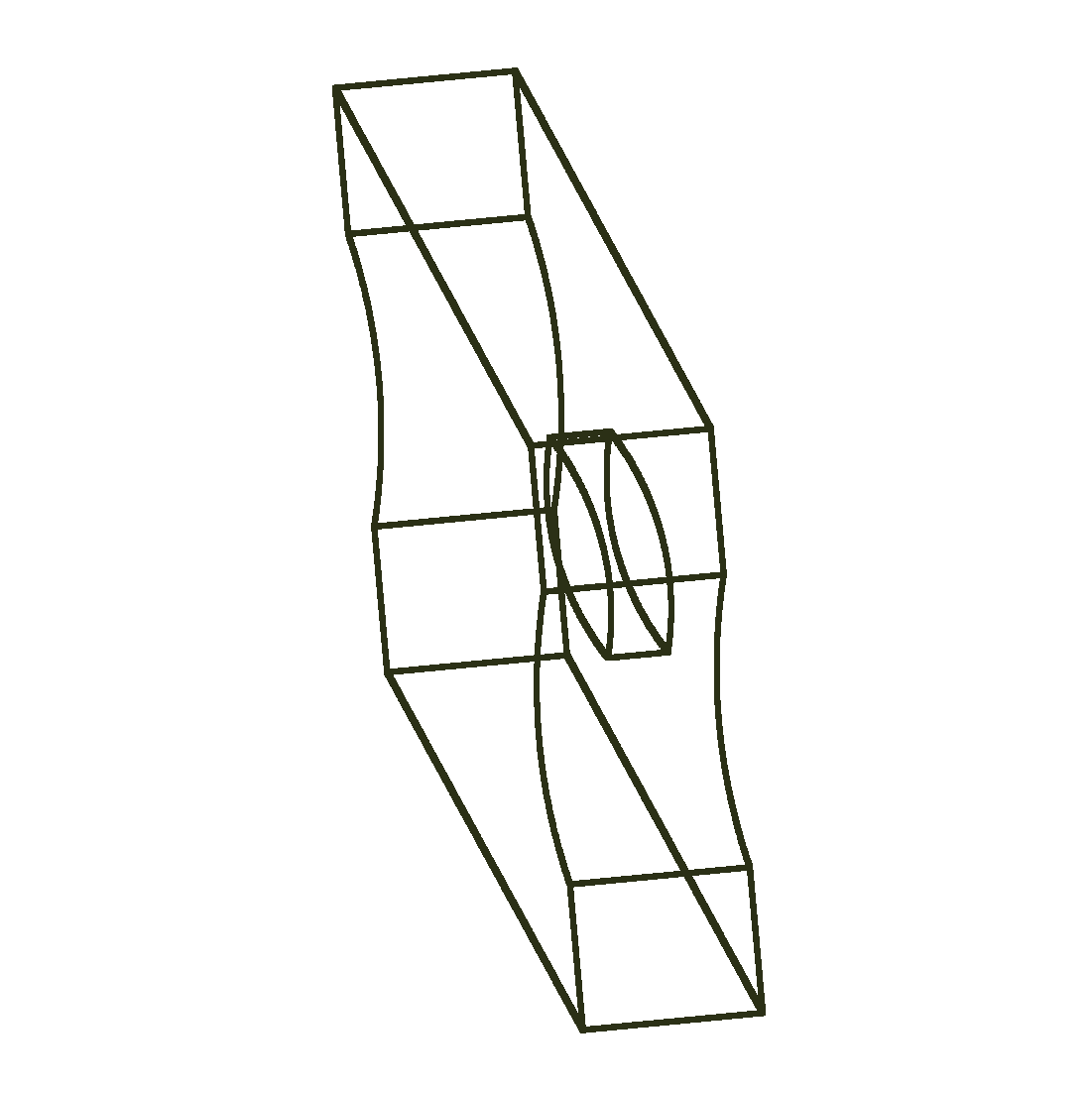} 
        \includegraphics[width=0.082\linewidth]{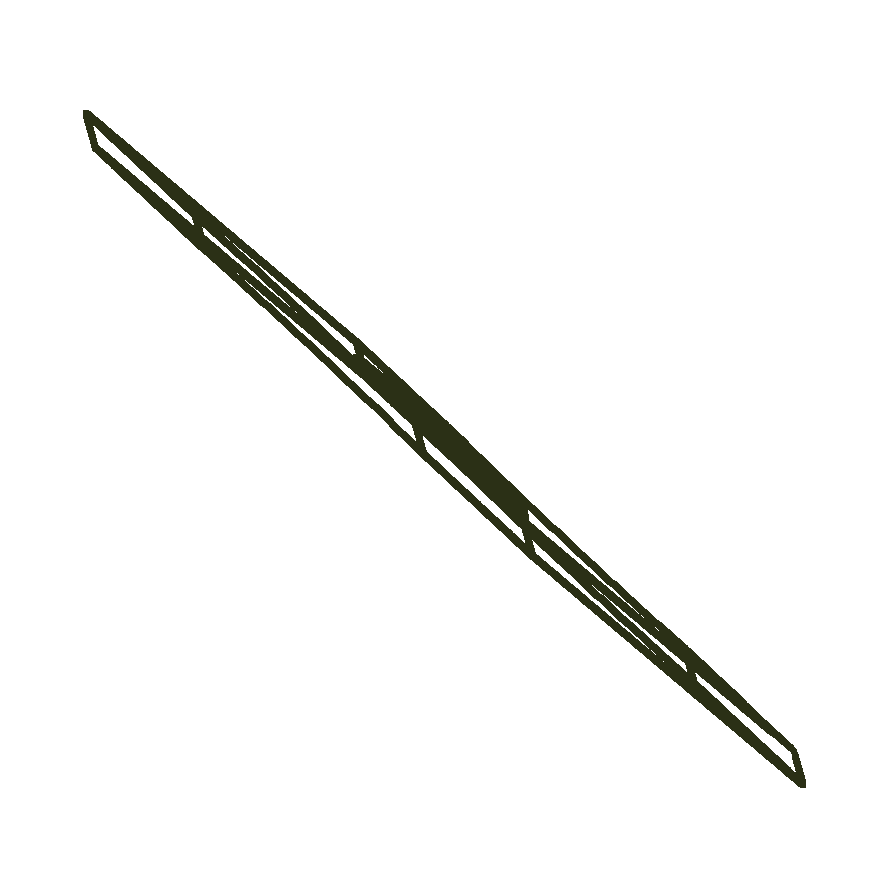}  
        \includegraphics[width=0.082\linewidth]{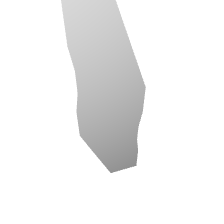} &
        \includegraphics[width=0.082\linewidth]{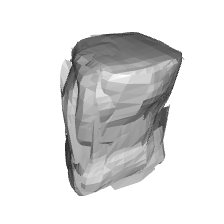} &

        \includegraphics[width=0.082\linewidth]{figures/reconstruction/input/00010156.png} &
        \includegraphics[width=0.082\linewidth]{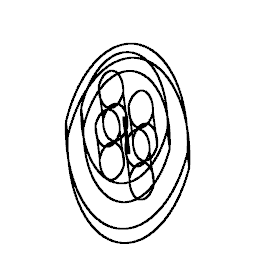} 
        \includegraphics[width=0.082\linewidth]{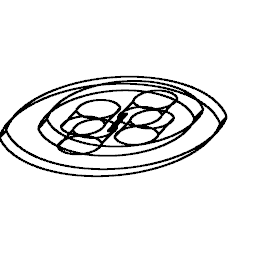}  
        \includegraphics[width=0.082\linewidth]{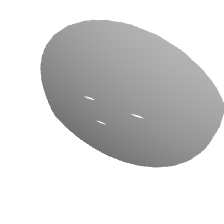} &
        \includegraphics[width=0.082\linewidth]{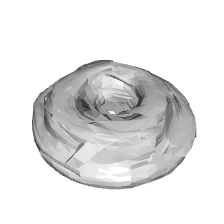} \tabularnewline

        \includegraphics[width=0.082\linewidth]{figures/reconstruction/input/00020864.png} &
        \includegraphics[width=0.082\linewidth]{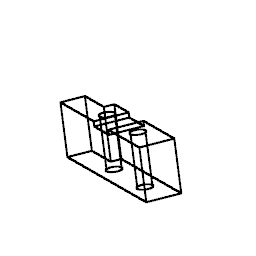} 
        \includegraphics[width=0.082\linewidth]{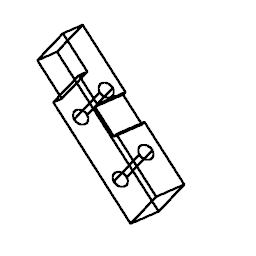}  
        \includegraphics[width=0.082\linewidth]{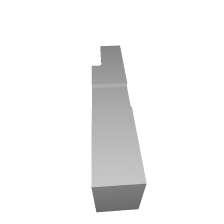} &
        \includegraphics[width=0.082\linewidth]{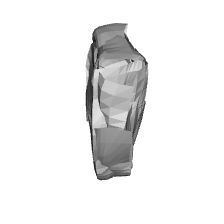} &

        \includegraphics[width=0.082\linewidth]{figures/reconstruction/input/00040184.png} &
        \includegraphics[width=0.082\linewidth]{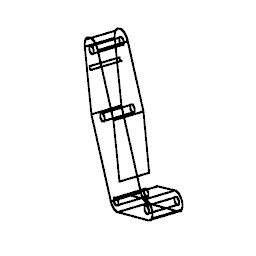} 
        \includegraphics[width=0.082\linewidth]{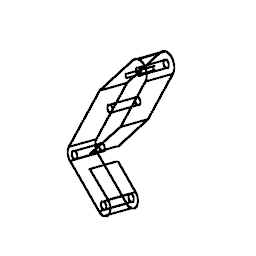}  
        \includegraphics[width=0.082\linewidth]{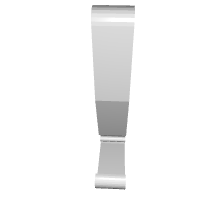} &
        \includegraphics[width=0.082\linewidth]{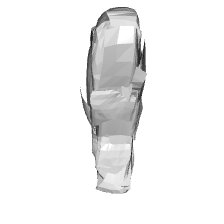} \tabularnewline
        \bottomrule
    \end{tabular}
    \caption{3D reconstruction results. For our method, we show two different views of the reconstructed 3D wireframe, plus the mesh from the same viewpoint as the input. For AtlasNet, we show the mesh from the same viewpoint as the input.}    
    \vspace{-3mm}
    \label{fig:reconstruction}
\end{figure*}

Figure~\ref{fig:reconstruction} shows 3D reconstruction results for various shapes using the predicted faces. In the \emph{first row}, we show two cases in which all faces are correctly identified and the 3D model is fully reconstructed. As a comparison, we also train AtlasNet~\cite{GroueixFKRA18}, a popular deep learning method for single-view 3D reconstruction, on our dataset and include the test results in Figure~\ref{fig:reconstruction}. Unlike our method, most existing deep learning methods take an image as input and generate unstructured point cloud or meshes in an end-to-end fashion. As one can see, such a method struggles to learn with the wireframe inputs because the features are very sparse when treated as an image.
In contrast, we propose to detect structures (\ie, face topology) in the wireframe and use geometric reasoning for 3D inference. Therefore, our method is able to output clean, structured 3D models.

The \emph{second row} and \emph{third row} of Figure~\ref{fig:reconstruction} show examples in which our face identification model makes some incorrect predictions (see Figure~\ref{fig:face-results}) but the 3D reconstruction results are unaffected. For the two cases in the second row, our model generates edge loops which are not part of the true face topology, but all included edges fall onto the same surface (\eg, a plane). For the two cases in the third row, the incorrect faces are filtered because they align to more than two dominant directions. Since the number of constraints created by the faces is typically larger than the number of unknowns in Eq.~\eqref{eqn:3d}, the 3D model can be recovered even if some faces are missed.

The \emph{fourth row} shows two examples in which incorrectly predicted faces affect local part of the 3D model. And the \emph{fifth row} shows two examples in which 3D reconstruction completely fails and the recovered shapes appear flat. The latter typically occurs when our model makes multiple mistakes (\eg, grouping edges on the opposite sides of the 3D shape into one face). Again, we refer readers to Figure~\ref{fig:face-results} for visualization of the mistakes made by our model.

In the \emph{last row} of Figure~\ref{fig:reconstruction}, we show two interesting cases in which even humans have trouble inferring the 3D geometry due to the chosen viewpoints. In contrast, our method does a decent job using the extracted face loops. This suggests that our method is not sensitive to the viewpoints.

\begin{figure}[t]
    \centering
    \includegraphics[width=3.1in]{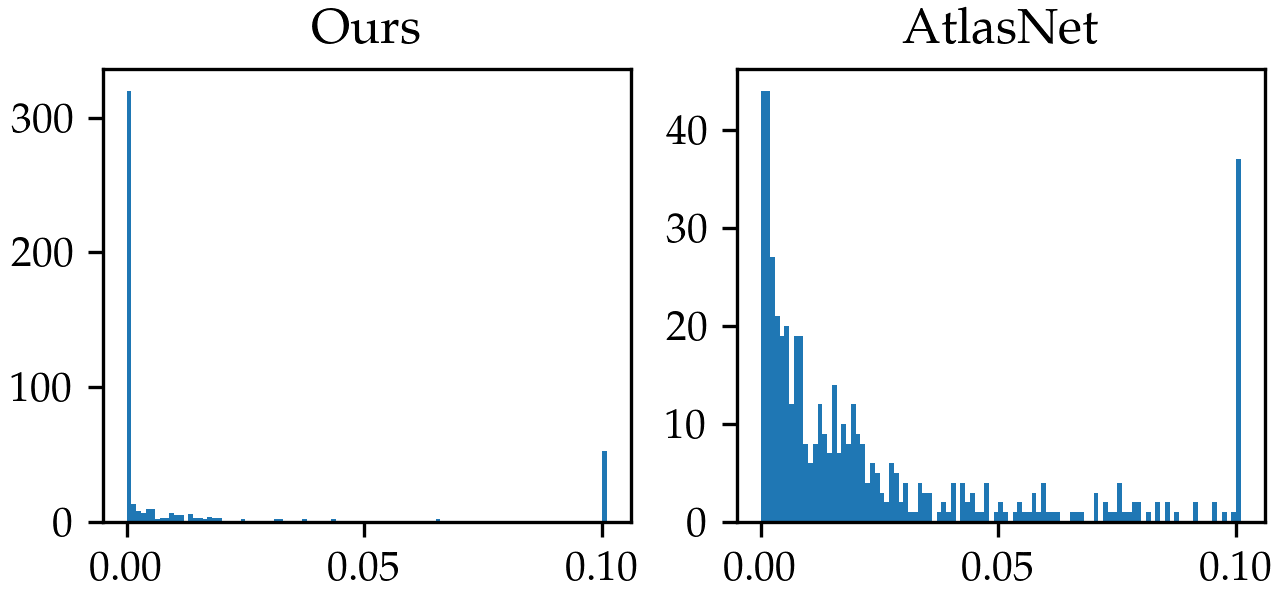}
    \caption{Histogram of Chamfer distances between our predicted meshes and the ground-truth meshes on 504 objects in test set.}
    \vspace{-3mm}
    \label{fig:chamfer}
\end{figure}

\smallskip
\noindent{\bf Quantitative results.} We compute the Chamfer distance between the predicted meshes and the ground truth in the test set, and plot the distributions in Figure~\ref{fig:chamfer}. As one can see, the structured 3D models obtained by our method tend to be more accurate. For example, for more than $60\%$ of the objects, our method achieves a distance of $<10^{-3}$.

\section{Discussion}

\noindent{\bf Limitations.} In this work, we present a data-driven approach to face identification. We point out that the proposed method should not be treated as a replacement or competitor to the geometry- and topology-based methods. Instead, we have found that our method complements existing techniques in several aspects, such as capturing the intent of designers, handling curved surfaces and disjoint components in the edge-vertex graph. Meanwhile, our model could make wrong predictions, while geometry- and topology-based methods are guaranteed to succeed when all assumptions are met.

Our method assumes that input 2D wireframe is clean and noise-free. In a practical system (\eg, SMARTPAPER~\cite{SheshC04}), this may be the outcome of multiple sketch cleaning and beautification steps. We hypothesize that our deep model can be made robust to noisy inputs with proper training (\eg, data augmentation), but a thorough investigation is beyond the scope of this paper.

\smallskip
\noindent{\bf Future directions.} Our work opens up several directions for future work. One direction we are particularly interested in is identifying conflicting constraints (due to incorrect predictions made by a deep network) in a geometric constraint system for 3D reconstruction. In this work, we treat all predicted faces equally. A better solution will not only improve the 3D reconstruction results, but may also generalize to other types of structural constraints in CAD.

{\small
\bibliographystyle{ieee_fullname}
\bibliography{cvpr22-face}
}

\end{document}